\documentclass[letterpaper, 10 pt, journal, twoside]{IEEEtran}

\usepackage{times}
\usepackage{xifthen}
\usepackage{xparse}
\usepackage{leftidx}
\usepackage{bm}
\usepackage{etoolbox}
\usepackage{amsmath}
\usepackage{multirow, makecell}
\usepackage{xspace}

\usepackage[nolist,nohyperlinks]{acronym}

\newcommand{\bbm}{\begin{bmatrix}}
\newcommand{\ebm}{\end{bmatrix}}

\DeclareMathAlphabet{\mybf}{OT1}{ptm}{b}{n} % letters
\newcommand{\mybs}[1]{{\bm{#1}}} % symbols
\DeclareMathAlphabet{\mybfi}{OML}{cmm}{b}{it}

\newcommand{\mbf}[1]{
\ifcat\noexpand#1\relax % check if the argument is a control sequence
\mybs{#1}% probably Greek
\else
\mybf{#1}% single character
\fi
}

\newcommand{\mbfbar}[1]{{\overline{\mbf{#1}}}}
\newcommand{\mbfhat}[1]{{\hat{\mbf{#1}}}}
\newcommand{\mbftilde}[1]{{\tilde{\mbf{#1}}}}

\newcommand{\mbfdot}[1]{{\dot {\mbf{#1}}}}

\NewDocumentCommand{\mbfidentity}{o}{\IfValueTF{#1}{\mbf{I}_{#1\hspace{\rightshift}}}{\mbf{I}}}
\NewDocumentCommand{\mbfzero}{oo}{\IfValueTF{#1}{\mbf{0}_{#1\times#2\hspace{\rightshift}}}{\mbf{0}}}

\newcommand{\cframe}[1]{{\smash{\protect\underrightarrow{\mathcal{F}}_{#1}}}}

\newlength{\leftshift}
\newlength{\rightshift}
\newcommand{\pos}[2]{\leftidx{_{#1}}{ \mbf r}{_{#2\hspace{\rightshift}}}} % position
\newcommand{\posbar}[2]{\leftidx{_{#1}}{\mbfbar r}{_{#2\hspace{\rightshift}}}} % position
\newcommand{\lm}[1]{\leftidx{_{#1}}{\mbf l}} % landmark
\NewDocumentCommand{\vel}{moo}{
	\IfValueTF{#1}{\leftidx{_{#1}}}{}{\mbf v}{\IfValueTF{#2}{_{#2#3\hspace{\rightshift}}}{}}}
\NewDocumentCommand{\veltilde}{moo}{
	\IfValueTF{#1}{\leftidx{_{#1}}}{}{\mbftilde v}{\IfValueTF{#2}{_{#2#3\hspace{\rightshift}}}{}}}
\NewDocumentCommand{\velbar}{moo}{
	\IfValueTF{#1}{\leftidx{_{#1}}}{}{\mbfbar v}{\IfValueTF{#2}{_{#2#3\hspace{\rightshift}}}{}}}
\NewDocumentCommand{\velhat}{moo}{
	\IfValueTF{#1}{\leftidx{_{#1}}}{}{\mbfhat v}{\IfValueTF{#2}{_{#2#3\hspace{\rightshift}}}{}}}
\NewDocumentCommand{\veldot}{moo}{
	\IfValueTF{#1}{\leftidx{_{#1}}}{}{\mbfdot v}{\IfValueTF{#2}{_{#2#3\hspace{\rightshift}}}{}}}

\NewDocumentCommand{\acc}{moo}{
	\IfValueTF{#1}{\leftidx{_{#1}}}{}{\mbf a}{\IfValueTF{#2}{_{#2#3\hspace{\rightshift}}}{}}}
\NewDocumentCommand{\acctilde}{moo}{
	\IfValueTF{#1}{\leftidx{_{#1}}}{}{\mbftilde a}{\IfValueTF{#2}{_{#2#3\hspace{\rightshift}}}{}}}
\NewDocumentCommand{\accbar}{moo}{
	\IfValueTF{#1}{\leftidx{_{#1}}}{}{\mbfbar a}{\IfValueTF{#2}{_{#2#3\hspace{\rightshift}}}{}}}
\NewDocumentCommand{\acchat}{moo}{
	\IfValueTF{#1}{\leftidx{_{#1}}}{}{\mbfhat a}{\IfValueTF{#2}{_{#2#3\hspace{\rightshift}}}{}}}
\NewDocumentCommand{\accdot}{moo}{
	\IfValueTF{#1}{\leftidx{_{#1}}}{}{\mbfdot a}{\IfValueTF{#2}{_{#2#3\hspace{\rightshift}}}{}}}

\NewDocumentCommand{\rotvel}{moo}{
	\IfValueTF{#1}{\leftidx{_{#1}}}{}{\mbf $\omega$}{\IfValueTF{#2}{_{#2#3\hspace{\rightshift}}}{}}}
\NewDocumentCommand{\rotveltilde}{moo}{
	\IfValueTF{#1}{\leftidx{_{#1}}}{}{\mbftilde $\omega$}{\IfValueTF{#2}{_{#2#3\hspace{\rightshift}}}{}}}
\NewDocumentCommand{\rotvelbar}{moo}{
	\IfValueTF{#1}{\leftidx{_{#1}}}{}{\mbfbar $\omega$}{\IfValueTF{#2}{_{#2#3\hspace{\rightshift}}}{}}}
\NewDocumentCommand{\rotvelhat}{moo}{
	\IfValueTF{#1}{\leftidx{_{#1}}}{}{\mbfhat $\omega$}{\IfValueTF{#2}{_{#2#3\hspace{\rightshift}}}{}}}
\NewDocumentCommand{\rotveldot}{moo}{
	\IfValueTF{#1}{\leftidx{_{#1}}}{}{\mbfdot $\omega$}{\IfValueTF{#2}{_{#2#3\hspace{\rightshift}}}{}}}

\newcommand{\C}[2]{ {\mbf C}   {_{#1#2\hspace{\rightshift}} }     } % rotation matrix
\newcommand{\T}[2]{{\boldsymbol{T}}{_{#1#2\hspace{\rightshift}}}} % homogeneous transformation matrix
\newcommand{\q}[2]{{\mbf q}{_{#1#2\hspace{\rightshift}}}} % quaternion of rotation
\newcommand{\qbar}[2]{{\mbfbar q}{_{#1#2\hspace{\rightshift}}}} % quaternion of rotation
\newcommand{\SEthree}{SE(3)}
\newcommand{\SOthree}{SO(3)}

\begin{acronym}[AAAAAAAAA]
    \acro{1d}[1D]{One-Dimensional}
    \acro{2d}[2D]{Two-Dimensional}
    \acro{2fast}[2Fast-2Lamaa]{Fast Field-based Agent-Subtracted Tightly-coupled Lidar Localisation And Mapping with Accelerometer and Angular-rate}
    \acro{3d}[3D]{Three-Dimensional}
    \acro{ate}[ATE]{Absolute Trajectory Error}
    \acro{ba}[BA]{Bundle Adjustment}
    \acro{cas}[CAS]{Centre for Autonomous Systems}
    \acro{cnn}[CNN]{Convolutional Neural Network}
    \acro{cpu}[CPU]{Central Processing Unit}
    \acro{doals}[DOALS]{Urban Dynamic Objects LiDAR}
    \acro{dof}[DoF]{Degree-of-Freedom}
    \acro{dvs}[DVS]{Dynamic Vision Sensor}
    \acrodefplural{dvs}[DVS's]{Dynamic Vision Sensors}
    \acro{ekf}[EKF]{Extended Kalman filter}
    \acro{fmcw}[FMCW]{Frequency Modulated Continuous Wave}
    \acro{fifo}[FIFO]{First In, First Out}
    \acro{fov}[FoV]{Field-of-View}
    \acro{gnss}[GNSS]{Global Navigation Satellite System}
    \acrodefplural{gnss}[GNSS's]{Global Navigation Satellite Systems}
    \acro{gp}[GP]{Gaussian Process}
    \acrodefplural{gp}[GPs]{Gaussian Processes}
    \acro{gpm}[GPM]{Gaussian Preintegrated Measurement}
    \acro{ugpm}[UGPM]{Unified Gaussian Preintegrated Measurement}
    \acro{gps}[GPS]{Global Position System}
    \acrodefplural{gps}[GPS's]{Global Position Systems}
    \acro{gpu}[GPU]{Graphic Processing Unit}
    \acro{hdr}[HDR]{High Dynamic Range}
    \acro{hri}[HRI]{Human-Robot Interactions}
    \acro{icp}[ICP]{Iterative Closest Point}
    \acro{iid}[i.i.d.]{independent and identically distributed}
    \acro{idol}[IDOL]{IMU-DVS Odometry using Lines}
    \acro{iou}[IoU]{Intersection over Union}
    \acro{imu}[IMU]{Inertial Measurement Unit}
    \acro{in2laama}[IN2LAAMA]{INertial Lidar Localisation Autocalibration And MApping}
    \acro{kf}[KF]{Kalman Filter}
    \acro{kl}[KL]{Kullback–Leibler}
    \acro{lidar}[LiDAR]{Light Detection And Ranging Sensor}
    \acro{lpm}[LPM]{Linear Preintegrated Measurement}
    \acro{map}[MAP]{Maximum A Posteriori}
    \acro{mle}[MLE]{Maximum Likelihood Estimation}
    \acro{ndt}[NDT]{Normal Distribution Transform}
    \acro{pca}[PCA]{Principal Component Analysis}
    \acro{pm}[PM]{Preintegrated Measurement}
    \acro{rcnn}[R-CNN]{Region-based Convolutional Neural Network}
    \acro{rrbt}[RRBT]{Rapidly Exploring Random Belief Trees}
    \acro{rgb}[RGB]{color}
    \acro{rgbd}[RGBD]{color-depth}
    \acro{rms}[RMS]{Root Mean Squared}
    \acro{rmse}[RMSE]{Root Mean Squared Error}
    \acro{ros}[ROS]{Robot Operating System}
    \acro{sde}[SDE]{Stochastic Differential Equation}
    \acro{SE3}[SE(3)]{Special Euclidean group in three dimensions}
    \acro{slam}[SLAM]{Simultaneous Localisation And Mapping}
    \acro{SO3}[SO(3)]{Special Orthonormal rotation group in three dimensions}
    \acro{tsdf}[TSDF]{Truncated Signed Distance Field}
    \acro{upm}[UPM]{Upsampled-Preintegrated-Measurement}
    \acro{uts}[UTS]{University of Technology, Sydney}
    \acro{vi}[VI]{Visual-Inertial}
    \acro{vio}[VIO]{Visual-Inertial Odometry}
    \acro{vo}[VO]{Visual Odometry}
\end{acronym}
\usepackage{graphicx}

\usepackage{fancyhdr}

\fancypagestyle{firstpage}{
  \fancyhf{}
  \fancyhead[C]{\footnotesize This work has been submitted to the IEEE for possible publication.\\Copyright may be transferred without notice, after which this version may no longer be accessible.}
}
\usepackage{multicol}
\usepackage[bookmarks=true]{hyperref}
\usepackage[capitalise]{cleveref}
\crefname{equation}{}{}
\usepackage{xcolor}
\usepackage{tabularx}
\usepackage{booktabs}
\usepackage{multirow}
\usepackage{rotating}
\usepackage{hhline}
\usepackage{amssymb}
\usepackage{adjustbox}
\usepackage[table]{xcolor}

\newcolumntype{Y}{>{\centering\arraybackslash}X}

\begin{document}

\title{Scalix: Uncertainty-Aware Scale-Consistent\\Monocular SLAM}

% You will get a Paper-ID when submitting a pdf file to the conference system
\author{Sebastian Barbas Laina*, Tianyi Zhang*, Panagiotis Petropoulakis, Simon Schaefer, Simon Boche, Jaehyung Jung, \\ Cedric Le Gentil, Stefan Leutenegger}
%\author{Author Names Omitted for Anonymous Review.}

%\author{\authorblockN{Michael Shell}
%\authorblockA{School of Electrical and\\Computer Engineering\\
%Georgia Institute of Technology\\
%Atlanta, Georgia 30332--0250\\
%Email: mshell@ece.gatech.edu}
%\and
%\authorblockA{Twentieth Century Fox\\
%\and
%\authorblockN{James Kirk\\ and Montgomery Scott}
%\authorblockA{Starfleet Academy\\
%San Francisco, California 96678-2391\\

% avoiding spaces at the end of the author lines is not a problem with
% conference papers because we don't use \thanks or \IEEEmembership

%Homer Simpson\authorrefmark{2},
%Montgomery Scott\authorrefmark{3} and
%Eldon Tyrell\authorrefmark{4}}
%Telephone: (800) 555--1212, Fax: (888) 555--1212}
%\authorblockA{\authorrefmark{4}Tyrell Inc., 123 Replicant Street, Los Angeles, California 90210--4321}}

\maketitle
\IEEEpeerreviewmaketitle
\thispagestyle{firstpage}
\pagestyle{fancy}

\begin{abstract}
Cameras are ubiquitous sensors in robotics due to their compact form factor and the perceptual richness captured through visual information. Monocular SLAM enables robots to understand the environment with a minimum setup, however, it inherently suffers from scale ambiguity. A common solution is to provide multi-modal sensor configurations, such as visual–inertial systems, where scale is observable unless the robot navigates under a constant-velocity motion, a common scenario in mobile robotics. With the advent of deep-learning, geometric foundation models have been used to address this problem, but the depths maps are often noisy and scale-inconsistent across frames.
In this paper, we propose Scalix, a real-time monocular SLAM framework that achieves metric-scale state estimation by integrating learned depth cues into a probabilistic factor-graph formulation. By augmenting existing monocular depth models with both per-pixel depth uncertainty and per-frame scale uncertainty, Scalix treats scale predictions as independent measurements within its optimization, leading to improved scale consistency through multi-view data associations. Experiments in large-scale outdoor and indoor environments demonstrate state-of-the-art performance on both metric and up-to-scale benchmarks while maintaining real-time operation and generalization. The code will be released upon acceptance.
\end{abstract}

\section{Introduction}
% first try for push
\ac{slam} is at the core of many robotic applications, such as scene understanding and 3D reconstruction.
Cameras have become the sensor of choice due to their low cost and ability to capture rich texture information of the environment.
Nonetheless, a single camera is scale-ambiguous, making it unable to recover the metric or absolute scale of the environment. Multi-camera or multi-modal sensor configurations \cite{campos2021orb, leutenegger2022okvis2, boche2025okvis2, geneva2020openvins} can address this problem, but require accurate extrinsic calibration, which is time-consuming and error-prone.
Additionally, the form-factor of certian embodiments (e.g. small unnmaned aerial vehicles) only allows for monocular-inertial configurations, where the IMU becomes the only source of scale information. Nonetheless, the scales becomes non-observable under constant-velocity motions, motivating the need of metric-scale monocular SLAM for improved robustness. 
At the same time, monocular \ac{slam} enables state estimation from casual online videos, thereby producing  metric-scale annotated data required by downstream robot learning tasks \cite{chen2025vidbot, zhang2025actron3d}.

Pioneered by traditional pipelines such as \cite{campos2021orb, gao2018ldso}, recent monocular \ac{slam} works have achieved impressive results by using learnt priors such as: monocular depth \cite{tateno2017cnn, zhu2024nicer}, dense optical flow \cite{teed2021droid, lipson2024deep}, and multi-view geometry \cite{murai2025mast3r, wang2025continuous, deng2025vggt, zhang2025vista}. However, these methods are limited by the scale inconsistencies of the learned models, causing scale drift or impeding metric-scale estimation.

\begin{figure}[t]
  \centering
  \includegraphics[width=\linewidth]{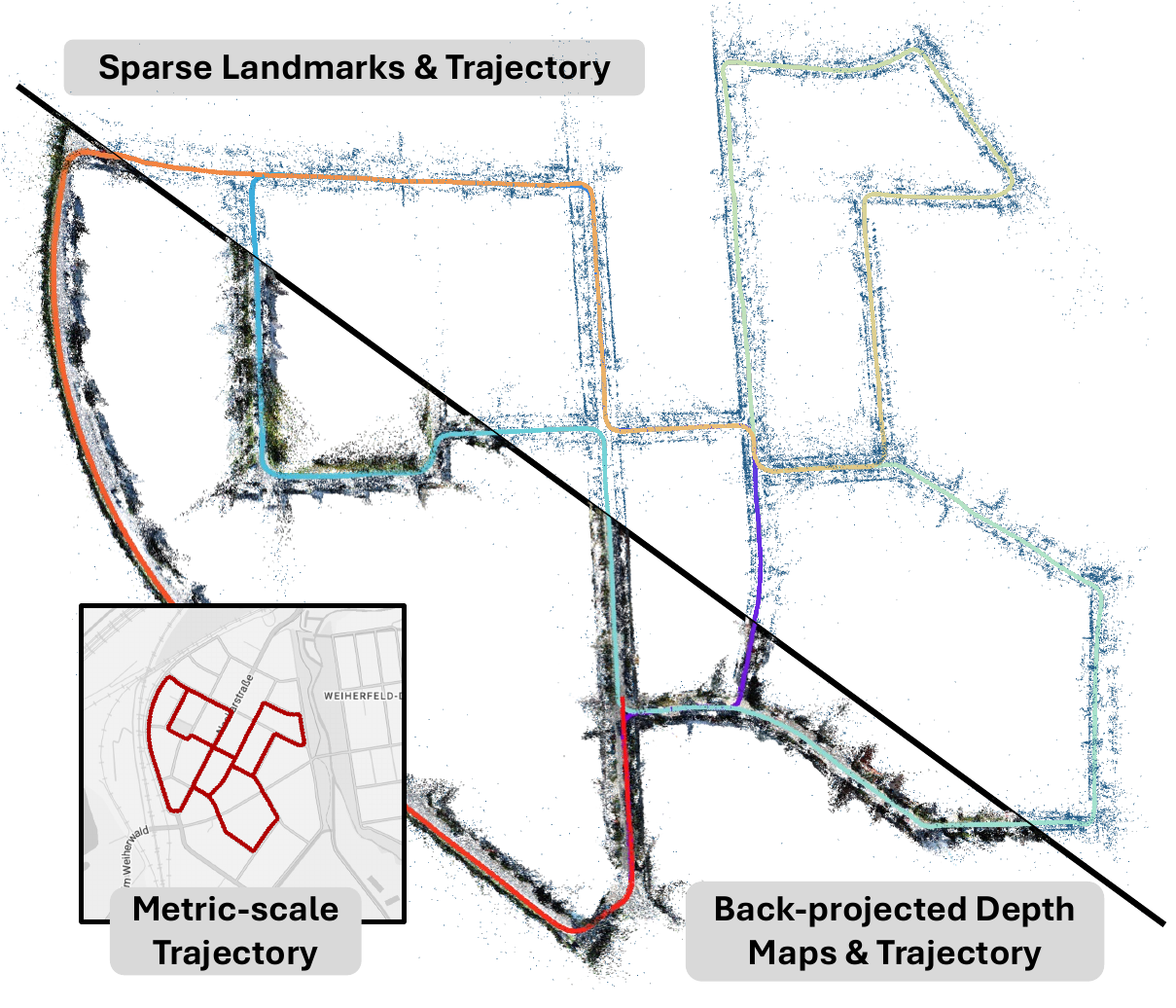}
  \caption{\textbf{Scalix on a Large-scale Sequence \cite{Geiger2012CVPR}.} 
  Scalix achieves accurate, metric-scale trajectory estimation on challenging scenes. 
  The main figure shows the estimated trajectory (colored by time), the sparse landmarks (upper-right diagonal) and the dense back-projected depth maps scaled using the optimized scale factors (lower-left diagonal). 
  In the lower-left corner, we directly align the trajectory to the map via SE(3) alignment.
  }
  \label{fig:teaser}
  \vspace{-0.1cm}
\end{figure}

In this paper, we present Scalix, a monocular \ac{slam} system that tightly couples visual data with learned scale and depth priors, enhancing scale consistency in the estimation process.
%, an example of our metric-scale estimation can be seen in \cref{fig:teaser}.
By incorporating a scale into a factor-graph formulation, Scalix diminishes the scale ambiguities between consecutive frames via multi-view data associations, improving the overall estimated trajectory.
This is done in a fully probabilistic fashion by learning the uncertainties from the latent representations of foundation monocular depth models.
It allows the method to be deployed in a zero-shot fashion in different scenarios, including large-scale environments as illustrated in \cref{fig:teaser}.

% \begin{figure}[t]
%   \centering
%   \includegraphics[width=\linewidth]{figures/teaser/teaser_sparse.pdf}
%   \caption{\textbf{Scalix Results on a Large-scale Sequence \cite{Geiger2012CVPR}.} Scalix achieves accurate, metric-scale trajectory estimation on challenging scenes while running in real-time. The main figure shows the estimated trajectory (colored by time) and the sparse point cloud. In the lower-left corner, we directly align the trajectory to the map via SE(3) alignment.}
% \end{figure}

The main contributions of this paper are:
\begin{itemize}
    \item A novel state-estimation backend parameterization that treats the depth image's scales as optimized variables constrained by multi-view data association and unary priors in a factor-graph framework with marginalization for efficient real-time performance on CPU.
    \item A metric depth decoupling formulation that models the per-frame scale error for depth-from-monocular foundation models, enabling the prediction of both the per-pixel depth uncertainty and scale uncertainty based on the latent representation of existing models.
    \item The integration of the previous concepts into Scalix, a real-time metric-scale monocular SLAM method that achieves state-of-the-art results.
    %\item We introduce Scalix, a real-time metric-scale monocular \ac{slam} method that achieves state-of-the-art results by leveraging monocular metric depth prediction.
    %\item We propose a novel parametrization of the state-estimation backend that treats the scale of a depth image as a measurement to be optimized in a tightly-coupled fashion, enabling the correction of the network's predicted depth. Furthermore, we incorporate this scale parameter into the marginalization process for real-time performance on CPU devices.
    %\item We propose a metric depth decoupling formulation to model the per-frame scale error. Our method learns and predicts the decoupled per-pixel depth uncertainty and per-frame scale uncertainty, based on existing monocular depth foundation models.
    % \item The formulation enables multi-sensor fusion, allowing for other modalities (e.g. IMU) to also improve the estimated states and correct the depth measurements provided by the camera.
\end{itemize}

\section{Related Work}
\label{sec:related_work}
\subsection{Monocular Visual \ac{slam}}
Monocular visual \ac{slam} addresses the problem of recovering camera trajectory and scene structure using a single freely moving camera as the sole data source.
Classic methods \cite{engel2017direct, engel2014lsd, davison2007monoslam, campos2021orb, klein2007parallel, forster2014svo} achieve this by minimizing image observation errors, including photometric errors, landmark reprojection errors, or a combination of both.
Subsequent approaches \cite{teed2021droid, lipson2024deep} introduce learning-based features to improve matching quality, such as dense optical flow or deep patches, further enhancing robustness and overall performance.
To achieve real-time performance and keep the optimization problem bounded, most methods adopt sliding window bundle adjustment.
However, relying exclusively on image observations, monocular visual \ac{slam} systems suffer from scale ambiguity, making absolute metric scale unobservable.
In addition to the global scale ambiguity, scale drift accumulates over large-scale trajectories, further limiting the applicability of monocular \ac{slam} systems in real-world scenarios.

\subsection{Monocular \ac{slam} with Depth Priors}
% 1. introuduction of Monocular depth models
Recent developments in monocular metric depth prediction \cite{bhat2023zoedepth, hu2024metric3d, piccinelli2025unidepthv2, yang2024depthv2, wang2025moge2} have enabled models to predict dense 3D depth in real-world metrics from a single RGB image by inherently capturing semantic information and incorporating camera intrinsics, demonstrating generalization ability across domains. 
While monocular metric depth prediction provides valuable depth priors for monocular \ac{slam}, temporal inconsistency across frames limits its direct applicability.
To address this issue, several \ac{slam} pipelines \cite{zhang2025hi, sandstrom2025splat, zhu2024nicer} leverage monocular depth priors for initialization and couple depth errors with image observation errors in bundle adjustment to improve geometric consistency and trajectory quality.
However, these dense methods rely on either computationally expensive dense bundle adjustment or iterative optimization, sacrificing real-time performance.

With the rise of generic transformer architectures \cite{caron2021emerging, oquab2023dinov2}, recent 3D foundation models \cite{wang2024dust3r, wang2025vggt, keetha2025mapanything} have demonstrated the ability to directly regress more consistent 3D point maps, poses, and camera parameters by jointly processing multiple images in an end-to-end manner.
Following this direction, feedforward SLAM methods \cite{murai2025mast3r, wang2024spann3r, wang2025continuous, zhang2025vista, maggio2025vggt, deng2025vggt, chen2026lingbot, zhang2026loger} extend these ideas to sequential input.
Several works utilize 3D foundation models in the frontend, either by leveraging correspondences \cite{murai2025mast3r} or employing frame chunks and local submaps \cite{zhang2025vista, maggio2025vggt, deng2025vggt}, followed by a classic optimization backend. However, these methods still suffer from scale drift between chunks, and some require a full chunk upfront rather than processing frames sequentially, preventing them from being fully streamable. 
Another line of work, such as Spann3R \cite{wang2024spann3r}, CUT3R \cite{wang2025continuous}, and subsequent methods \cite{chen2026lingbot, zhang2026loger}, addresses this limitation by maintaining an internal implicit memory through attention mechanisms. However, these approaches suffer from forgetting issues, experience scale drift in long-term scenarios, and their GPU requirements make them difficult to deploy in onboard or resource-constrained applications.

\begin{figure*}[t]
  \centering
  % Using RGB (0–255)
  \definecolor{mygrey}{HTML}{D1D1D1}
  \definecolor{mypink}{HTML}{EBD8F7}
  \includegraphics[width=0.95\linewidth]{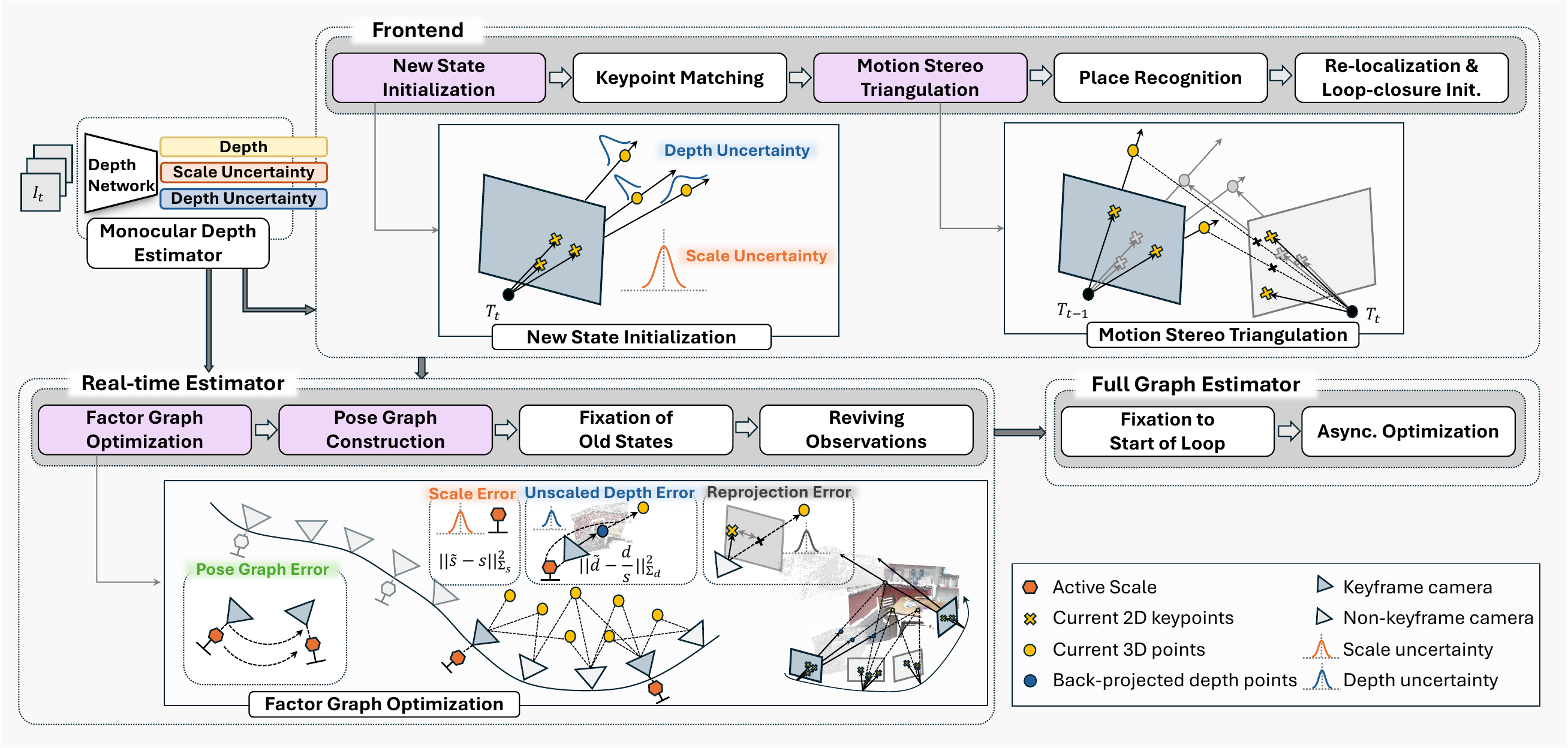}
  \caption{\textbf{Scalix System Overview.} 
  Our method builds upon OKVIS2 \cite{leutenegger2022okvis2}, a classical SLAM architecture comprising a frontend and a backend.
  Components with {\setlength{\fboxsep}{1pt}\colorbox{mypink}{pink background}} indicate our key contributions.
  Each keyframe is associated with an additional scale state \protect\adjustbox{valign=c}{\includegraphics[height=\fontcharht\font`A]{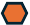}} and we use an enhanced monocular depth model to obtain depth and scale measurements. 
  In the frontend, landmark positions are initialized using the predicted depth and scale and the backend adds depth and scale error terms into the factor graph optimization. The pose graph is extended to account for the scale in the marginalization process.  }
  \label{fig:pipeline}
  \vspace{-0.3cm}
\end{figure*}

\section{Notation and Overview}
\label{sec:notation}

%%%% main notations
\NewDocumentCommand{\state}{O{}}{\mbf{x}^{#1}}
\NewDocumentCommand{\var}{O{}}{\mbf{\Sigma}_{#1}}
\NewDocumentCommand{\weight}{O{} O{}}{\mbf{W}^{#2}_{#1}}
% scale
\NewDocumentCommand{\s}{O{}}{s^{#1}}
\NewDocumentCommand{\smean}{O{}}{\Tilde{s}^{#1}}
% depth
\NewDocumentCommand{\depthvec}{O{} O{}}{\mbf{d}^{#2}_{#1}}
\NewDocumentCommand{\depthvecmean}{O{}}{\Tilde{\mbf{d}}^{#1}}
% \NewDocumentCommand{\depth}{O{} O{}}{d^{#2}_{#1}}
\newcommand{\depth}{d}
\NewDocumentCommand{\depthmean}{O{}}{\Tilde{\mathrm{d}}^{#1}}
% error
\NewDocumentCommand{\err}{O{} O{}}{e^{#2}_{#1}}
\NewDocumentCommand{\errvec}{O{} O{}}{\mbf{e}^{#2}_{#1}}
%others
\NewDocumentCommand{\hessian}{O{} O{}}{\mbf{H}^{#2}_{#1}}
\NewDocumentCommand{\jacobian}{O{} O{}}{\mbf{E}^{#2}_{#1}}
\NewDocumentCommand{\gradvec}{O{} O{}}{\mbf{b}^{#2}_{#1}}
\newcommand{\projfunc}{\pi}
%%%% tmp notations and subscripts
% variables & functions
\newcommand{\sigmavec}{\boldsymbol{\sigma}}
\newcommand{\img}{\mbf{I}}
\newcommand{\param}{\mbf{\theta}}
\newcommand{\pix}{\mbf{u}}
\newcommand{\posegraph}{\mbf{p}}
\newcommand{\cauchy}{\rho_{\mathrm{c}}}
\newcommand{\unitvec}{\mbf{n}}
\newcommand{\cost}{g}
% descriptions
\newcommand{\reproj}{\text{r}}
\newcommand{\depthDesc}{\text{d}}
\newcommand{\scaleDesc}{\text{s}}
\newcommand{\posegraphDesc}{\text{p}}
\newcommand{\refFrame}{r}
\newcommand{\conFrame}{c}

% Classical monocular \ac{slam} pipelines jointly estimate camera poses and scene structure by minimizing feature matching errors. 
% In monocular systems, the absence of absolute scale leads to accumulated scale drift over time. 
Scalix proposes a tightly coupled, scale-aware monocular \ac{slam} system that addresses scale inconsistencies by incorporating the depth and scale measurements from a monocular depth model. 
We propose a learning formulation that jointly models and trains the scale and depth uncertainties of a foundation depth network (\cref{sec:depth_decoupling}), allowing the scale to be used as a measurement that can be explicitly modeled as an optimizable variable by our \ac{slam} pipeline, presented in \cref{fig:pipeline}. For convenience, we first define the notations used here.

In the monocular visual \ac{slam} problem, a moving camera is tracked with respect to a fixed world reference frame $\cframe{W}$. We denote the camera coordinate frame by $\cframe{C}$. 
The rigid body transformation $\T{A}{B} \in \SEthree$ transforms homogeneous points between two frames: $\pos{A}{P} = \T{A}{B}\pos{B}{P}$, where $\pos{A}{P}$ is the position of a point $P$ in frame $\cframe{A}$. The rotational part of $\T{A}{B}$ is expressed by $\C{A}{B} \in \SOthree$ and $\pos{A}{B} \in \mathbb{R}^3$ denotes the translation component. 
We also denote the rotation $\C{A}{B}$ with its unit quaternion form $\q{A}{B}$. 
The covariance matrix of a variable $\mbf{a}$ is represented as $\var[\mbf{a}]$. In Scalix, each new image at time step $k$ is associated with a state, represented as:
\begin{equation}
\label{eq:state-vector}
    \state[] = \left[
    \pos{W}{C}^{\top}, 
    \q{W}{C}^{\top}, 
    \vel{W}^{\top}, 
    \s[]
    \right]^{\top},
\end{equation}
where $\pos{W}{C}$, $\q{W}{C}$ and $\vel{W}$ denote the position, orientation, and velocity of the camera in $\cframe{W}$, and $\s[]$ is the estimated depth scale factor. 

\section{Metric Depth Decoupling}
\label{sec:depth_decoupling}

\subsection{Formulation}
Our monocular metric depth network takes a single image $\img$ as input to obtain geometrically plausible depth maps.
Rather than purely local pixel-wise noise, the dominant failure mode often manifests as inaccurate global scale prediction and scale drift with respect to real-world geometry.
To better characterize this behavior, we reformulate the relationship between the predicted depth $\depthvec$ and the true metric depth $\depthvec[\text{m}]$ through a metric depth decoupling formulation:
\begin{equation}
\label{eq:depth_decouple}
    \depthvec[\text{m}] = \s \, \depthvec ,
\end{equation}
where $\s$ is a global depth scale factor for each image.
Under the assumption of a metric depth prediction model that gives depths close to the true metric values, we assume that the scale follows a normal distribution $\s \sim \mathcal{N}(\smean, \Sigma_{\s})$ with mean $\smean = 1$ and variance $\Sigma_{\s}=\sigma_{\s}^2$.
Apart from the global scale uncertainty, the depth prediction also exhibits pixel-wise independent uncertainty. 
We assume that the unscaled depth $\depthvec$ follows a normal distribution $\depthvec \sim \mathcal{N}(\depthvecmean, \var[\depthvec])$, with a predicted mean depth $\Tilde{\depthvec}$ and a variance $\var[\depthvec] = \mathrm{diag}(\sigmavec_{\depthvec}^2)$, where both $\sigmavec_{\depthvec}$ and $\Tilde{\depthvec}$ denote the vectorized per-pixel values.
This probabilistic decoupling formulation enables subsequent optimization within a unified probabilistic framework that eliminates the need for iterative scale alignment between frames or submaps during graph optimization.

\subsection{Decoupled Prediction in Monocular Depth Model}
\begin{figure}[t]
  \definecolor{mydarkblue}{HTML}{BFC9D5}
  \definecolor{myyellow}{HTML}{FDF4D0}
  \definecolor{mypurple}{HTML}{E0D6E6}
  \centering
    \includegraphics[width=.96\linewidth]{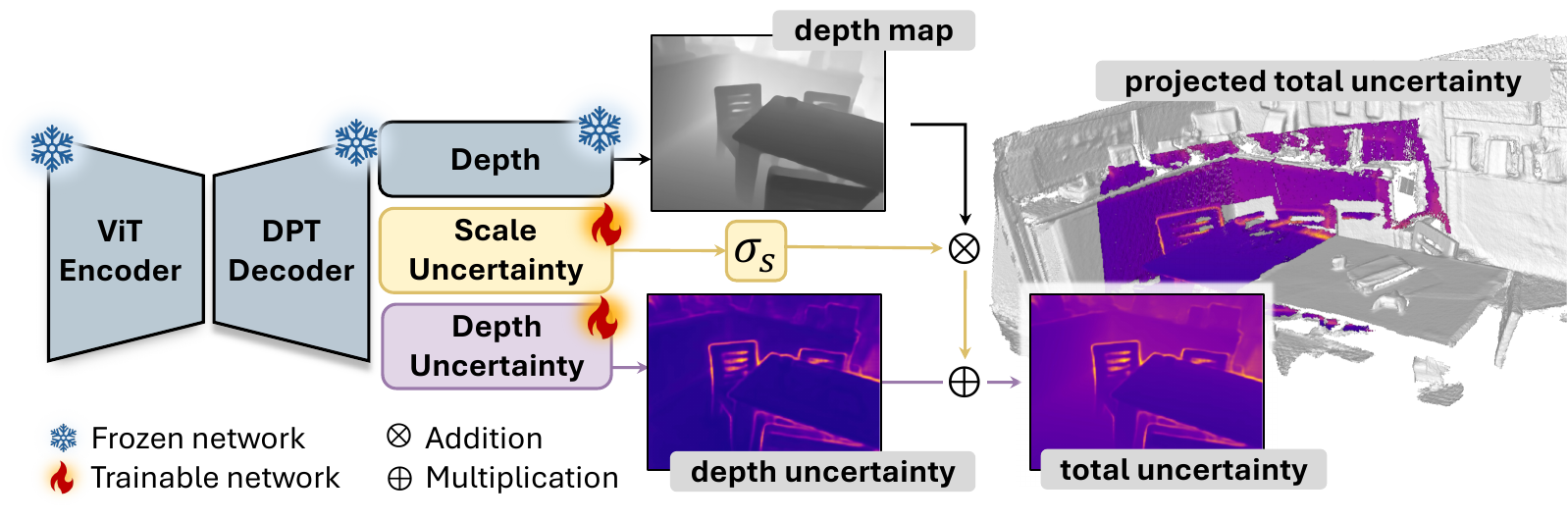}
  \caption{\textbf{Monocular Depth Network Augmented with Decoupled Uncertainty Prediction.}
  Modules with {\setlength{\fboxsep}{1pt}\colorbox{mydarkblue}{dark blue background}} denote the original depth prediction network architecture in Metric3Dv2 \cite{hu2024metric3d}. The {\setlength{\fboxsep}{1pt}\colorbox{myyellow}{yellow box}} denotes the scale uncertainty head, which outputs a global per-frame scale as a 1D scalar.
  The {\setlength{\fboxsep}{1pt}\colorbox{mypurple}{purple box}} denotes the depth uncertainty head, which outputs independent per-pixel uncertainty.
  The two uncertainties are combined to form the total uncertainty, and we visualize the projected total uncertainty in 3D space.
}
\vspace{-0.2cm}
  \label{fig:uncertainty}
\end{figure}

In most metric depth prediction networks, only depth and pixel-wise independent confidence are provided, and the global depth scale is assumed to be $1$.
To predict both the global scale uncertainty and the independent depth uncertainty described above, we augment a base monocular depth model with an uncertainty decoder head and adopt a probabilistic loss for learning aleatoric uncertainty. As shown in \cref{fig:uncertainty},
we define two uncertainty prediction networks, $f_{\s}$ for global scale uncertainty and $f_{\depth}$ for pixel-wise independent uncertainty:
\begin{equation}
\label{eq:prediction_network}
\begin{aligned}
\Sigma_{\s} &= f_{\s}(\img), \\
\var[\depthvec] &= f_{\depthvec}(\img).
\end{aligned}
\end{equation}

During training, supervision is available only in the form of metric depth measurements.
Therefore, both sources of uncertainty must be expressed in the metric depth domain during training. 
Since the scale is applied globally to all pixels, all pixel-wise depths are correlated, resulting in non-zero cross-correlation terms in the total metric depth uncertainty. 
Consequently, the metric depth uncertainty is not independent across pixels. 
Based on \eqref{eq:depth_decouple}, we derive the metric depth covariance matrix $\var[\depthvec[\text{m}]]$ via uncertainty propagation
\begin{equation}
    \label{eq:uncertainty_propagation}
    \var[\depthvec[\text{m}]] = \Sigma_{\s} \depthvec \depthvec[][\top] + {\s}^2\var[\depthvec].
\end{equation}

We then adopt the multivariate Gaussian negative log-likelihood loss to learn uncertainty:
\begin{equation}
\label{eq:NLL_matrix}
    \mathcal{L}(\param) = (\s\depthvec - \depthvec[\text{m}, \text{gt}])^{\top} \var[\depthvec[\text{m}]]^{-1}(\s\depthvec - \depthvec[\text{m}, \text{gt}]) + \mathrm{log}|\var[\depthvec[\text{m}]]|,
\end{equation}
where $\param$ denotes the network weights and $\depthvec[\text{m}, \text{gt}]$ denotes the ground truth value. 
Directly computing \eqref{eq:NLL_matrix} would require large-scale matrix operations. We further derive the element-wise formulation by applying Sherman–Morrison formula~\cite{sherman1950adjustment}: 
\begin{equation}
\label{eq:NLL_element}
\begin{aligned}
\mathcal{L}(\param) =& \frac{1}{{\s}^2} \sum_{i=1}^{N}\frac{{\err[i][2]}}{\sigma_{\depth_{i}}^2} - \lambda \left(\sum_{i=1}^{N}\frac{\err[i] \depth_i}{\sigma_{\depth_{i}}^2}\right)^2 + N\log ({s}^2) \\
& + \sum_{i=1}^{N}\log(\sigma_{\depth_i}^2) + 
\log\left(1 + \frac{\sigma_{\s}^2}{{\s}^2}\sum_{i=1}^N \frac{{\depth_i}^2}{\sigma_{\depth_i}^2}\right),
\end{aligned}
\end{equation}
where $\depth_i$ and $\sigma_{\depth_i}$ denote the depth and independent uncertainty of $i$-th pixel, respectively, and $\err[i] = \s \depth_i - \depth_{\text{m}, \text{gt}, i}$ is the error at $i$-th pixel. 
The scalar $\lambda$ is defined as $\lambda = \frac{\sigma_{\s}^2}{{\s}^4} / \left(1 + \frac{\sigma_{\s}^2}{{\s}^2} \sum_{i=1}^N \frac{{\depth_i}^2}{\sigma_{\depth_i}^2}\right)$.

Our approach is agnostic to the network architecture, although we conducted our experiments using Metric3Dv2 \cite{hu2024metric3d}. 
Metric3Dv2 adapts ViT~\cite{oquab2023dinov2} and DPT~\cite{ranftl2021vision} as en-decoder backbone to produce a hidden feature map and a confidence map, and then employs a CNN-based depth head enhanced with a GRU-based recurrent architecture to predict the final pixel-wise depth, normal and confidence. 
Our uncertainty prediction head $f_{\depth}$ adopts the same CNN architecture as the depth head in Metric3Dv2, and takes the feature map, the initial depth map, and the depth confidence map as input. 
The scale uncertainty head $f_{\s}$ takes the feature map for global context information and adopts a series of strided convolutions followed by adaptive average pooling and an MLP to aggregate spatial features into a one-dimensional output.

\section{Scale-aware Monocular \ac{slam}}
\label{sec:slam}

We illustrate our \ac{slam} pipeline in \cref{fig:pipeline}.
We adopt OKVIS2 \cite{leutenegger2022okvis2} formulation of frames and keyframes and also utilize a real-time graph that optimizes a set of states and landmarks within a local window. 
Another full-graph is used to optimize the full trajectory upon loop-closure detections, enabling real-time local and global trajectory estimation in parallel. 
Place recognition is based on DBoW2 \cite{GalvezTRO12}, with additional geometric verification rules as originally proposed in \cite{leutenegger2022okvis2} to filter potentially incorrect loop closures.
We also use the same keypoint detection and description algorithm described in \cite{leutenegger2011brisk}. 

\subsection{Visual Frontend}
\label{sec:slam_frontend}

As shown in~\cref{fig:pipeline}, the visual frontend mainly performs states and landmarks initialization. Camera poses are initialized through a 3D-2D matching step following~\cite{leutenegger2022okvis2}. The following describes the landmark initialization process for a monocular camera in detail.
For each keyframe, new landmarks are initialized using the predicted depth from the monocular depth network described in~\cref{sec:depth_decoupling} together with the associated scale:
\begin{equation}
\lm{C_k}{^j} = \s \depthvec[][][\pix] \projfunc^{-1}(\pix), \quad \lm{W}{^j} = \T{W}{C_k}\lm{C_k}{^j},
\end{equation}
where $\lm{C_k}{^j}$ and $\lm{W}{^j}$ denotes the $j$-th landmark in $k$-th keyframe and world coordinate respectively, $\pix$ is the pixel location of the keypoint, $\projfunc^{-1}(\cdot)$ is the inverse projection function, and $\depthvec[][][\pix]$ is the predicted depth at pixel location $\pix$. We only initialize landmarks with predicted depths within the valid range $\depth \in [0.01\text{m}, 30\text{m}]$.

In addition to the depth-based initialization in keyframes, our frontend also initializes landmarks through triangulation between frames and earlier co-visible keyframes. In this case, depth measurements that were previously rejected due to violating the valid depth range can be incorporated later if they pass a $\chi^2$ consistency test with a significance level of $\beta = 0.05$, computed as 
\begin{equation}
\frac{(\T{C_k}{W} \, \lm{W}{^j} - \s_k \depthvec[k][][\pix])^2}{\Sigma_{\depthvec[\text{m}_k][][\pix]}} < \beta.
\end{equation}
In this case, $C_k$ denotes the keyframe at timestamp $k$ involved in the triangulation, and $\Sigma_{\depthvec[\text{m}_k][][\pix]}$ is the propagated metric-depth uncertainty of keypoint $\pix$, computed according to~\eqref{eq:uncertainty_propagation}.
By combining these two strategies, our frontend achieves robust landmark initialization by leveraging both monocular depth predictions and multi-view geometric observations.

\subsection{Scale-aware Backend Optimization}
\label{sec:slam_backend}
For backend optimization, we use three error terms: the reprojection error of the landmarks, the depth error between a landmark's depth and the predicted depth, and the difference between the scale prior and the optimized scale. 

\subsubsection{Reprojection error}
We use the standard reprojection error $\errvec[\reproj][j, k]$ of the $j$-th
landmark $\lm{W}{^j}$ in the camera image at time step $k$:
\begin{equation}
\errvec[\reproj_{j, k}][] = \Tilde{\pix}_{j,k} - \projfunc(\T{C_k}{W} \, \lm{W}{^j}),
\end{equation}
with the detected keypoint pixel location being $\Tilde{\pix}_{j,k}$ and  $\projfunc(\cdot)$ denoting the camera projection. 
These errors constrain the rotation and the up-to-scale translation of a camera pose with respect to $\cframe{W}$, therefore requiring additional error terms to produce metric-scale pose estimations.  

\subsubsection{Depth error} 
We propose a depth error term that exploits the learned depth prior from monocular depth prediction networks, constraining 3D points via unscaled depth while enabling the optimization to refine our scale measurements:
\begin{equation}
\label{eq:depth_error}
\err[\depthDesc_{j, k}][] = \Tilde{\depth}_{j, k} - \frac{\unitvec_z^{\top} \, \T{C_k}{W} \, \lm{W}{^j}}{\s_k},
\end{equation}
with $\Tilde{\depth}_{j,k}$ being the network's predicted depth of a keypoint, $\s_k$ the estimated scale and $\unitvec_z$ is a constant vector $[0, 0, 1, 0]$ that extracts the z-coordinate of the landmark.

\subsubsection{Scale error} 
To leverage metric scale prediction, we anchor each frame's scale variable to a measurement with a prior error term:
\begin{equation}
\label{eq:scale_error}
\err[\scaleDesc_k][] = \smean_k - \s_k,
\end{equation}
with $\smean_k=1.0$ being the network's scale prediction and $\s_k$ the optimized scale in our problem.

\begin{figure}[t]
  \centering
  \includegraphics[width=0.97\linewidth]{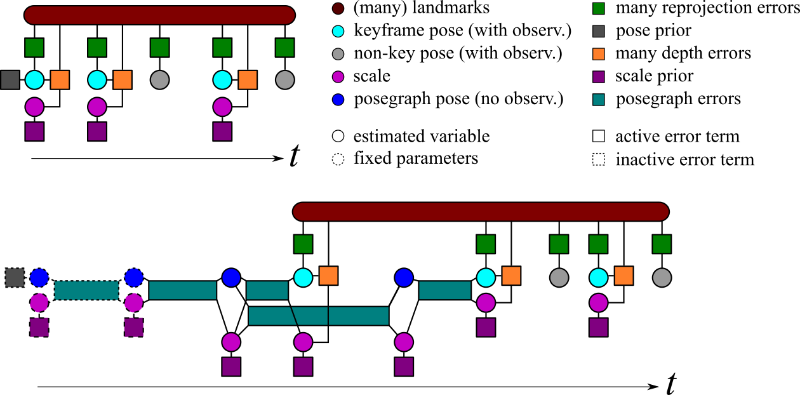}
  \caption{\textbf{Factor Graph in Scalix.} 
  Top: the real-time local window estimator. The current keyframe and non-keyframe states are connected by visual reprojection errors. 
  For every keyframe state in the optimization window, depth and scale errors are incorporated into the cost function. 
  Bottom: the full real-time estimator. Old measurements are marginalized and substituted by pose-scale graph edges, reducing the computational complexity.
  }
  \vspace{-0.2cm}
  \label{fig:graph}
\end{figure}

\subsection{Marginalization and Relative Pose–Scale Errors}
\label{sec:marginalization}
To ensure real-time operation, we marginalize old states to keep the optimization problem bounded. 
Following the strategy of~\cite{leutenegger2022okvis2}, our marginalization is performed between two keyframes $\refFrame, \conFrame$ by eliminating the co-observed landmarks. 
This process converts the associated state error terms into relative pose error terms, yielding a pose graph where the nodes represent pose states and the edges encode relative pose errors.
We refer the reader to~\cite{leutenegger2022okvis2} for a detailed explanation of how a Maximum Spanning Tree (MST) is employed to select the pose-graph edges to be created.
Compared to~\cite{leutenegger2022okvis2}, Scalix proposes to add scale into states:
\begin{equation}
\label{eq:posescale_state}
    \posegraph = [\pos{C_{\refFrame}}{C_{\conFrame}}, \q{C_{\refFrame}}{C_{\conFrame}}, \s_{\text{rel}}],
\end{equation}
where $\pos{C_{\refFrame}}{C_{\conFrame}}$ and $\q{C_{\refFrame}}{C_{\conFrame}}$ are the relative position and orientation expressed in the reference frame $\cframe{C_{\refFrame}}$, and $\s_{\text{rel}} = \frac{\s_{\conFrame}}{\s_{\refFrame}}$ denotes the relative scale. 
Therefore, the reprojection errors and depth errors with respect to the reference frame $\refFrame$ become
\begin{align}
\label{eq:posegraph_residual}
    \errvec[{\reproj_{j, \refFrame}}] = \Tilde{\pix}_{j,\refFrame} - \projfunc(\lm{C_{\refFrame}}{^j}), \, & 
    \errvec[{\reproj_{j, \conFrame}}] = \Tilde{\pix}_{j,\conFrame} - \projfunc(\T{C_{\conFrame}}{C_{\refFrame}} \, \lm{C_{\refFrame}}{^j}), \\
    \err[{\depthDesc_{j, \refFrame}}] = \Tilde{\depth}_{j, \refFrame} - \frac{\unitvec_z^{\top} \, \lm{C_{\refFrame}}{^j}}{\s_\refFrame}, \, &
    \err[{\depthDesc_{j, \conFrame}}] = \Tilde{\depth}_{j, \conFrame} - \frac{\unitvec_z^{\top} \, \T{C_{\conFrame}}{C_{\refFrame}} \, \lm{C_{\refFrame}}{^j}}{\s_\refFrame \, \s_{\text{rel}}},
\end{align}
where we treat $\s_\refFrame$ as constant in practice. Adopting landmark marginalization to all joint observations, we have the cost
\begin{equation}
\label{eq:posescale_cost}
\resizebox{\columnwidth}{!}{$\displaystyle
\begin{aligned}
    \cost_{\posegraphDesc_{\refFrame, \conFrame}} =
    & \frac{1}{2} \sum_{i=1}^{N} \sum_{j \in \mathcal{J}\left(\refFrame\right)} \cauchy \left( {\errvec[{\reproj_{j, \refFrame}}][\top]} \weight[\reproj] \errvec[{\reproj_{j, \refFrame}}][] \right)
    + \sum_{j \in \mathcal{S}\left(\refFrame\right)} \cauchy \left( {\err[{\depthDesc_{j, \refFrame}}][\top]} \weight[{\depthDesc_{j, \refFrame}}] \err[{\depthDesc_{j,\refFrame}}][] \right) \\
    & + \frac{1}{2} \sum_{i=1}^{N} \sum_{j \in \mathcal{J}\left(\conFrame\right)} \cauchy \left( {\errvec[{\reproj_{j, \conFrame}}][\top]} \weight[\reproj] \errvec[{\reproj_{j, \conFrame}}][] \right)
    + \sum_{j \in \mathcal{S}\left(\conFrame\right)} \cauchy \left( {\err[{\depthDesc_{j, \conFrame}}][\top]} \weight[\depthDesc_{j, \conFrame}] \err[{\depthDesc_{j, \conFrame}}][] \right),
\end{aligned}
$}
\end{equation}
with the sets $\mathcal{J}\left(\refFrame\right), \mathcal{J}\left(\conFrame\right)$ denoting the covisible landmarks at frames $C_\refFrame$ and $C_\conFrame$, with $\mathcal{S} \subset \mathcal{J}$ denoting the keypoints that have an associated depth term.
Now we approximate this cost as
\begin{align}
\label{eq:relative_posescale_error}
\cost_{\posegraphDesc_{\refFrame, \conFrame}} & = \frac{1}{2} {\errvec[\posegraphDesc_{r, c}][\top]} \weight[\posegraphDesc_{r,c}][] \errvec[\posegraphDesc_{r,c}][], \\
\errvec[\posegraphDesc_{\refFrame, \conFrame}][] & = \errvec[\posegraphDesc, 0_{\refFrame, \conFrame}][] +
    \begin{bmatrix}
    \pos{C_{\refFrame}}{C_{\conFrame}} - \posbar{C_{\refFrame}}{C_{\conFrame}} \\
    \q{C_{\refFrame}}{C_{\conFrame}} \boxminus \qbar{C_{\refFrame}}{C_{\conFrame}} \\
    \s_{\text{rel}} - \bar{\s}_{\text{rel}} \\
    \end{bmatrix},
\end{align}
where quantities with a bar (\,\(\bar{\cdot}\)\,) are the linearization points selected during marginalization, and $\errvec[\posegraphDesc, 0_{\refFrame, \conFrame}][]$ is a constant term computed through the marginalization process.
These terms are computed by applying Schur complement to the Gauss-Newton system that minimizes $\cost_{\posegraphDesc_{\refFrame, \conFrame}}$~\cref{eq:posescale_cost}:
\begin{equation}
\label{eq:marg_GN}
    \begin{bmatrix}
    \hessian[\posegraphDesc,\posegraphDesc]
    & \ldots & \hessian[\posegraphDesc, j] & \ldots\\
    \vdots & \ddots & \mbfzero & \mbfzero\\
    \hessian[\posegraphDesc,j][\top] & \mbfzero &\hessian[j,j] & \mbfzero\\
    \vdots & \mbfzero & \mbfzero &\ddots\\
    \end{bmatrix} 
    \begin{bmatrix}
    \delta \posegraph\\
    \vdots\\
    \delta \lm{}{^j} \\
    \vdots
    \end{bmatrix}
    = 
    \begin{bmatrix}
    \gradvec[\posegraphDesc] \\
    \vdots\\
    \gradvec[j]\\
    \vdots
    \end{bmatrix},
\end{equation}
with $\delta \lm{}{^j}$ the update of the $j$-th landmark in the reference frame $C_{\refFrame}$, and $\hessian[j, j]$, $\hessian[\posegraphDesc, j]$ the corresponding Hessian blocks associated with the $j$-th landmark for simplicity. 
Now, landmarks are marginalized out using the Schur complement:
\begin{equation}
\label{eq:marg_schur}
    \begin{aligned}
        \hessian[][*] &= \hessian[\posegraphDesc,\posegraphDesc] - \sum_{j}\hessian[\posegraphDesc,j] \hessian[j,j][\dagger] \hessian[\posegraphDesc,j][\top], \\
        \gradvec[][*] &= \gradvec[\posegraphDesc] - \sum_{j}\hessian[\posegraphDesc,j] \hessian[j,j][\dagger]\gradvec[j]. 
    \end{aligned}
\end{equation}
This yields the reduced Gauss-Newton system:
\begin{equation}
\label{eq:marg_GN_reduced}
\hessian[][*] 
\delta \posegraph 
= \gradvec[][*],
\end{equation}
where $\gradvec[][*]$ can be iteratively updated through $\gradvec[t][*] = \gradvec[][*] - \hessian[][*]\Delta{\mbf{\chi}}$ following \cite{leutenegger2022okvis2}, with $\Delta{\mbf{\chi}}$ denoting the state update term in \eqref{eq:relative_posescale_error}.
In this way, it is equivalent to a respective pose-scale error term of the following form
\begin{equation}
\label{eq:marg_optim}
{\jacobian[\posegraphDesc_{r,c}][\top]}\weight[\posegraphDesc_{r,c}][]\jacobian[\posegraphDesc_{r,c}][]\delta \posegraph = -{\jacobian[\posegraphDesc_{r,c}][\top]}\weight[\posegraphDesc_{r,c}][](\errvec[\posegraphDesc, 0_{\refFrame, \conFrame}][] + \Delta \mbf{\chi}),
\end{equation}
with
$
    {\jacobian[\posegraphDesc_{r, c}][\top]} = \mbfidentity[7], \weight[\posegraphDesc_{r,c}][] = \hessian[][*], \errvec[\posegraphDesc, 0_{r,c}][] =  \hessian[][*\dagger] \gradvec[][*]. 
$
Note that in our system, as stated in \cref{sec:slam_frontend}, there may be cases where a keyframe is not associated with a scale estimate or where some landmarks lack depth observations. 
If a frame lacks scale in the marginalization process, we set the linearization point of the relative scale to \(1\). When neither frame has scale association, the error term degenerates to the standard 6-dim relative pose error term described in~\cite{leutenegger2022okvis2}.

\subsection{Factor-Graph Optimization}
\label{sec:slam_optim}
All of the aforementioned factors are combined in the overall minimization objective:
\begin{align}
    \label{eq:optimization_objective}
    \cost\left( \state\right) &= 
    \frac{1}{2} \sum_{k \in \mathcal{K}} \sum_{j \in \mathcal{J} \left(k \right)} \cauchy \left( {\errvec[\reproj_{j,k}][\top]} \weight[\reproj] \errvec[\reproj_{j,k}][]   \right)
    \nonumber \\
    &+\frac{1}{2} \sum_{k \in \mathcal{K}} \sum_{j \in \mathcal{S} \left(k \right)} \cauchy \left( {\err[\depthDesc_{j,k}][\top]} \weight[\depthDesc_{j,k}] \err[\depthDesc_{j,k}][]   \right)
    \nonumber \\
    &+ \frac{1}{2} \sum_{k \in \mathcal{P} \cup \mathcal{K} \setminus f} 
    {\err[\scaleDesc_k][\top]} \weight[\scaleDesc_k][] \err[\scaleDesc_k][] 
    + \frac{1}{2} \sum_{r\in \mathcal{P}} \sum_{c\in \mathcal{C}\left(r\right)} {\errvec[\posegraphDesc_{r,c}][\top]} \weight[\posegraphDesc_{r,c}][] \errvec[\posegraphDesc_{r,c}][].  
\end{align}
Here, the set $\mathcal{K}$ contains the most recent frames as well as keyframes with observations of visible landmarks in $\mathcal{J} \left(k \right)$ and with $\mathcal{S} \subset\mathcal{J}$ the keypoints that have a depth term associated to them at timestamp $k$.
$\mathcal{P}$ contains all pose-scale graph frames and $f$ denotes the most current frame. 
$\mathcal{C}\left( \refFrame \right) \subset \mathcal{P}$ is the set of all pose-scale graph frames connected to a frame $C_{\refFrame}$. 
The Cauchy robustifier $\cauchy \left(\cdot\right)$ is used for reprojection errors and depth-errors. 
In this formulation, $\weight[\text{a}] = \var[\text{a}]^{-1}$, with $\text{a} \in \{\reproj, \depthDesc, \scaleDesc, \posegraphDesc\}$, represents the weight of each residual, defined as the inverse of its covariance matrix.
An example of our factor-graph formulation can be seen in \cref{fig:graph}.

\section{Experiments}
\label{sec:experiment}

%%%%%%%%%%%%%%%%%%% 6.3 update
\begin{table*}[t]
    \centering
    \caption{Trajectory evaluation on the Kitti dataset. RMSE ATE [m] with SIM(3) and SE(3) alignments (\textbf{Best}, and \underline{second best}).}
    \setlength{\tabcolsep}{1pt}
    \renewcommand{\arraystretch}{1.1}
    \footnotesize
    \def\failcolor{gray}
    \newcommand{\len}[1]{\scriptsize{\textit{#1\,km}}}
    \begin{tabularx}{\linewidth}{ll@{\hspace{1pt}}YYYYYYYYYYY!{\vrule}YY}
        \toprule
        & \multirow{2}{*}{\textbf{Method}} & \textbf{00} & \textbf{01} & \textbf{02} & \textbf{03} & \textbf{04} & \textbf{05} & \textbf{06} & \textbf{07} & \textbf{08} & \textbf{09} & \textbf{10} & \multirow{2}{*}{\textbf{Avg.}} & \textbf{Avg.}
        \\
        & & \len{3.7} & \len{2.5} & \len{5.1} & \len{0.6} & \len{0.4} & \len{2.2} & \len{1.2} & \len{0.6} & \len{3.2} & \len{1.7} & \len{0.9} & & \textbf{success}
        \\
        \hhline{---------------}
        \multirow{11}{*}{\begin{sideways}SIM(3)\end{sideways}}
        & ORB-SLAM3\textsuperscript{$\dagger$}\scriptsize{\cite{campos2021orb}} & 8.94 & \textcolor{\failcolor}{103.76} & 24.90 & \textbf{0.76} & \underline{0.79} & 5.33 & 12.87 & 2.55 & \textbf{4.45} & 8.39 & 6.57 & \underline{16.30} & \textcolor{\failcolor}{7.55} \\
        & LDSO\scriptsize{\cite{gao2018ldso}} & 9.32 & \underline{11.68} & 31.98 & 2.85 & 1.22 & \underline{5.10} & 13.55 & 2.96 & \textcolor{\failcolor}{129.00} & 21.64 & 17.36 & 22.42 & \textcolor{\failcolor}{11.77} \\
        & DROID-SLAM\scriptsize{\cite{teed2021droid}} & 92.10 & \textcolor{\failcolor}{344.60} & 107.60 & 2.38 & 1.00 & \textcolor{\failcolor}{118.50} & \textcolor{\failcolor}{62.47} & \textcolor{\failcolor}{21.78} & \textcolor{\failcolor}{161.60} & \textcolor{\failcolor}{72.32} & \textcolor{\failcolor}{118.70} & 100.28 & \textcolor{\failcolor}{50.77} \\
        & DPV-SLAM++\scriptsize{\cite{lipson2024deep}} & \underline{8.30} & 11.86 & 39.64 & 2.50 & \textbf{0.78} & 5.74 & 11.60 & \underline{1.52} & \textcolor{\failcolor}{110.90} & \textcolor{\failcolor}{76.70} & 13.70 & 25.75 & \textcolor{\failcolor}{10.63} \\
        & DROID+M3D\textsuperscript{$\dagger$} \scriptsize{\cite{teed2021droid,hu2024metric3d}} & 12.84 & \textcolor{\failcolor}{150.44} & \underline{20.03} & 2.08 & 0.83 & 7.78 & \underline{3.19} & \textcolor{\failcolor}{18.32} & 9.73 & \textbf{3.76} & \underline{4.83} & 21.26 & \textcolor{\failcolor}{8.34} \\
        % \hhline{~--------------}
        & CUT3R\scriptsize{\cite{wang2025continuous}} & \textcolor{\failcolor}{190.38} & \textcolor{\failcolor}{90.59} & \textcolor{\failcolor}{264.39} & \textcolor{\failcolor}{20.40} & 7.31 & \textcolor{\failcolor}{92.25} & \textcolor{\failcolor}{67.54} & \textcolor{\failcolor}{22.48} & \textcolor{\failcolor}{145.08} & \textcolor{\failcolor}{67.42} & \textcolor{\failcolor}{40.00} & 91.62 & \textcolor{\failcolor}{\underline{7.31}} \\
        & ViSTA-SLAM\textsuperscript{$\dagger$}\scriptsize{\cite{zhang2025vista}} & \textcolor{\failcolor}{160.90} & \textcolor{\failcolor}{414.40} & \textcolor{\failcolor}{257.30} & \textcolor{\failcolor}{39.32} & \textcolor{\failcolor}{25.85} & \textcolor{\failcolor}{110.63} & \textcolor{\failcolor}{58.68} & \textcolor{\failcolor}{51.99} & \textcolor{\failcolor}{215.60} & \textcolor{\failcolor}{154.60} & \textcolor{\failcolor}{88.24} & 143.41 & \textcolor{\failcolor}{/} \\
        & VGGT-Long\scriptsize{\cite{deng2025vggt}} & 8.67 & \textcolor{\failcolor}{121.20} & 32.08 & 6.12 & 4.23 & 8.31 & 5.34 & 4.63 & 53.10 & 41.99 & 18.37 & 27.64 & \textcolor{\failcolor}{18.28} \\
        & LoGeR*\scriptsize{\cite{zhang2026loger}} & 30.47 & 47.91 & 36.32 & 5.30 & 1.95 & 26.34 & 6.60 & 5.55 & 24.41 & 10.12 & 10.11 & 18.64 & 18.64 \\
        & Lingbot-map\textsuperscript{$\dagger$}\scriptsize{\cite{chen2026lingbot}} & 24.24 & \textcolor{\failcolor}{83.63} & 65.75 & 2.50 & 1.02 & \textcolor{\failcolor}{74.77} & 7.51 & 9.87 & 19.45 & 42.55 & 19.69 & 31.91 & \textcolor{\failcolor}{21.40} \\
        % \hhline{~--------------}
        & \cellcolor{gray!15}Ours (w/o BA) & \cellcolor{gray!15}\textbf{2.85} & \cellcolor{gray!15}\textbf{10.72} & \cellcolor{gray!15}\textbf{15.04} & \cellcolor{gray!15}\underline{1.90} & \cellcolor{gray!15}1.04 & \cellcolor{gray!15}\textbf{1.84} & \cellcolor{gray!15}\textbf{1.62} & \cellcolor{gray!15}\textbf{0.91} & \cellcolor{gray!15}\underline{7.30} & \cellcolor{gray!15}\underline{4.76} & \cellcolor{gray!15}\textbf{2.57} & \cellcolor{gray!15}\textbf{4.60} & \cellcolor{gray!15}\textbf{4.60} \\
        \hhline{---------------}
        \multirow{2}{*}{\begin{sideways}SE(3)\end{sideways}}
        & DROID+M3D\textsuperscript{$\dagger$}\scriptsize{\cite{teed2021droid,hu2024metric3d}} & \underline{12.88} & \textcolor{\failcolor}{\underline{430.75}} & \underline{20.14} & \underline{15.13} & \underline{9.93} & \underline{8.88} & \underline{9.25} & \underline{22.29} & \underline{13.40} & \underline{9.65} & \underline{9.41} & \underline{51.06} & \textcolor{\failcolor}{\underline{12.07}} \\
        & \cellcolor{gray!15}Ours (w/o BA) & \cellcolor{gray!15}\textbf{3.96} & \cellcolor{gray!15}\textbf{21.47} & \cellcolor{gray!15}\textbf{16.55} & \cellcolor{gray!15}\textbf{9.44} & \cellcolor{gray!15}\textbf{3.25} & \cellcolor{gray!15}\textbf{4.30} & \cellcolor{gray!15}\textbf{1.99} & \cellcolor{gray!15}\textbf{2.15} & \cellcolor{gray!15}\textbf{7.66} & \cellcolor{gray!15}\textbf{4.86} & \cellcolor{gray!15}\textbf{3.46} & \cellcolor{gray!15}\textbf{7.19} & \cellcolor{gray!15}\textbf{7.19} \\
        \bottomrule
    \end{tabularx}
    \renewcommand{\arraystretch}{1.0}
    \raggedright
    \vspace*{1pt}
    
    \scriptsize{
    \textcolor{\failcolor}{Gray}: failures (ATE $>$ 3\% of seq. length).
    \textbf{Avg. success}: average excluding failed sequences. 
    Methods with \textsuperscript{$\dagger$} are run by ourselves.
    }
    \label{tab:kitti_traj_eval}
\end{table*}

\subsection{Experiment Setup}
\label{sec:experiment_setup}
\subsubsection{Evaluation Dataset and Metrics}
\label{sec:experiment_setup_dataset}
We evaluate our method on large-scale outdoor benchmarks, inherently prone to scale drift, and indoor scenarios to demonstrate that Scalix can generalize to different domains. 
We report \ac{rmse} of \ac{ate} in the KITTI dataset~\cite{Geiger2012CVPR} and 7-Scenes~\cite{shotton2013scene} after Sim(3) alignment. 
For methods that produce metric-scale estimates, we additionally report the results after SE(3) alignment. 
% For results not reported in the original paper, we use the median over three runs.

\subsubsection{Implementation Details}
\label{sec:experiment_setup_impl}
For training, we initialize Metric3Dv2~\cite{hu2024metric3d} with its small model variant, which employs DINOv2-reg (ViT-S)~\cite{oquab2023dinov2} as encoder and a DPT~\cite{ranftl2021vision} decoder. 
During training, we freeze the network and train the uncertainty heads using ScanNet~\cite{dai2017scannet} and Waymo~\cite{sun2020scalability} for one day using a single NVIDIA L40s GPU.
For the SLAM pipeline, we set the sliding window size to 10. All experiments were conducted on a PC equipped with an Intel i7-13700 CPU and an NVIDIA RTX 3080 GPU.

\subsection{Results}
\label{sec:experiment_result}
\subsubsection{KITTI dataset}
\label{sec:experiment_result_kitti}

\begin{figure*}[t]
  \centering
  \includegraphics[width=0.95\linewidth, trim={0mm 8mm 0mm 0mm}]{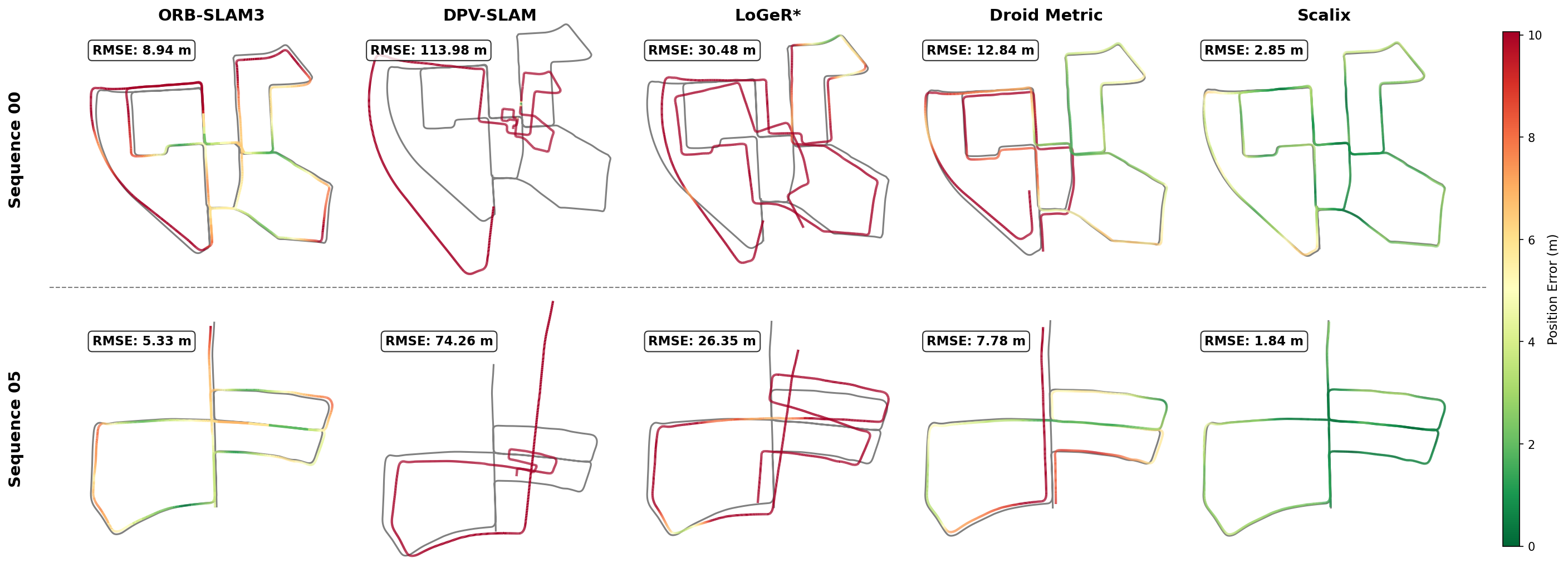}
  \caption{\textbf{Qualitative Trajectory Comparison on KITTI~\cite{Geiger2012CVPR} under Sim(3) Alignment.} 
  The color gradient from red to green indicates trajectory error from high to low. 
  The colored lines represent the estimated trajectories and the gray lines represent the ground-truth trajectories. 
}
\vspace{-0.2cm}
  \label{fig:traj_kitti}
\end{figure*}

The KITTI dataset~\cite{Geiger2012CVPR} provides grayscale and RGB stereo cameras, a LiDAR, and an IMU.
For our experiments, we only use the left camera from the RGB pair.
% We evaluate two versions of the network: \textit{ours small} refers to Scalix with the \textit{DINO2reg-ViT-Small} version from Metric3Dv2, while \textit{Ours big} is for the DINO2reg-ViT-Large version, showcasing the effect of having a better depth network, at the expense of higher runtime. 
Quantitative results are presented in~\cref{tab:kitti_traj_eval}, and qualitative results are shown in~\cref{fig:traj_kitti}, where Scalix outperforms other monocular methods.

Under Sim(3) alignment, our method achieves the best result among the optimization-based and feedforward methods, improving by $37\%$ over CUT3R\cite{wang2025continuous}, the second-best method.
Additionally, compared to DROID-SLAM+Metric3D~\cite{teed2021droid,hu2024metric3d}, which also leverages metric depth to improve odometry accuracy, our method achieves significantly better results, demonstrating the effectiveness of our pipeline integration.
On the KITTI dataset, some methods exhibit extremely large errors on certain sequences.
For a fair comparison, we treat trajectories with \ac{ate} greater than $3\%$ as failures and compute the average accuracy only over successful trajectories, reported in the ``Avg. success" column.
Even after filtering out failure cases, our method still performs best on average. 
Under SE(3) alignment, to our knowledge, only DROID+M3D (DROID-SLAM~\cite{teed2021droid} with Metric3D~\cite{hu2024metric3d}) achieves metric-scale reconstruction.
Using the same Metric3D model, our pipeline outperforms it by $40.4\%$.
Our results reported in \cref{tab:kitti_traj_eval} are obtained without full bundle adjustment (w/o BA). When full bundle adjustment is enabled, our method improves further, achieving an \ac{ate} of 
$3.87$m and $7.22$m under Sim(3) and SE(3) alignment. It is important to note that across our three runs, we observed similar results, except for sequence 1, where we observed one failure.

\subsubsection{7-scenes dataset}
\label{sec:experiment_result_7scenes}

The 7-Scenes dataset~\cite{shotton2013scene} consists of handheld recordings with an RGB-D Kinect camera in indoor scenarios. Our setup only uses the RGB data.
We compare our method with other baselines in~\cref{tab:7scenes_traj_eval}.
In indoor scenarios, Scalix still achieves on-par results with state-of-the-art methods, which can be further improved by adding the full BA component, achieving $7.8$cm and $11.5$cm \ac{ate} under Sim(3) and SE(3) alignment, respectively.
DROID-SLAM~\cite{teed2021droid}, ViSTA-SLAM~\cite{zhang2025vista} and MASt3R-SLAM~\cite{murai2025mast3r} achieve slightly better performance in indoor sequences. 
However, as shown in~\cref{tab:kitti_traj_eval}, these methods do not generalize well to large-scale outdoor scenes, either failing completely or exhibiting significant errors (Note that MASt3R-SLAM~\cite{murai2025mast3r} is omitted from \cref{tab:kitti_traj_eval} due to tracking loss in all sequences).
In contrast, our method maintains strong performance across both indoor and outdoor environments. 
The remaining gap in indoor sequences is due to the accuracy of Metric3Dv2~\cite{hu2024metric3d} in indoor scenes. 
This limitation is further evidenced by the performance degradation observed in DROID+M3D~\cite{teed2021droid, hu2024metric3d} compared to the original DROID-SLAM.
Compared to LoGeR*~\cite{zhang2026loger}, Lingbot-map~\cite{chen2026lingbot} and DROID+M3D~\cite{teed2021droid,hu2024metric3d}, which achieve decent results in both settings, our method achieves competitive results while also recovering metric-scale trajectories.
\subsubsection{Runtime analysis}
The system has two parallel threads: one for the frontend processing (keypoint matching, monocular inference and landmark initialization) and a second one for the backend optimization. Keypoint matching takes 20 ms, monocular network inference 79 ms (only triggered on a subset of frames), and landmark initialization 5 ms. The backend optimization is performed in 59ms.

\begin{table}[]
    \centering
    \caption{Trajectory evaluation on the 7scenes dataset. RMSE ATE [cm] with SIM(3) and SE(3) alignments (\textbf{Best}, and \underline{second best}).}
    \setlength{\tabcolsep}{1.5pt}
    \renewcommand{\arraystretch}{1.1}
    \footnotesize
    \def\failcolor{gray}
    \begin{tabularx}{\columnwidth}{ll@{\hspace{1pt}}YYYYYYY!{\vrule}Y}
        \toprule
        & \scriptsize{\textbf{Method}} & \scriptsize{\textbf{chess}} & \scriptsize{\textbf{fire}} & \scriptsize{\textbf{heads}} & \scriptsize{\textbf{office}} & \scriptsize{\textbf{pump.}} & \scriptsize{\textbf{kit.}} & \scriptsize{\textbf{stairs}} & \scriptsize{\textbf{Avg.}}
        \\
        \hhline{----------}
        \multirow{11}{*}{\begin{sideways}SIM(3)\end{sideways}}
        & \scriptsize{ORB-SLAM3\textsuperscript{$\dagger$}\cite{campos2021orb}} & 4.2 & 2.9 & 16.3 & 8.6 & 40.8 & 37.3 & \underline{1.2} & 15.9 \\
        & \scriptsize{LDSO\textsuperscript{$\dagger$}\cite{gao2018ldso}} & \underline{4.0} & 12.9 & 51.0 & 14.9 & -- & 49.9 & 23.9 & 26.1 \\
        & \scriptsize{DROID-SLAM\cite{teed2021droid}} & \textbf{3.6} & \underline{2.7} & \underline{2.5} & \underline{6.6} & \underline{12.7} & \textbf{3.5} & 2.6 & \underline{5.0} \\
        & \scriptsize{DROID+M3D\textsuperscript{$\dagger$}\cite{teed2021droid,hu2024metric3d}} & 4.9 & 6.4 & 9.6 & 14.2 & 16.3 & 4.7 & 14.6 & 10.1 \\
        & \scriptsize{CUT3R\cite{wang2025continuous}} & 5.9 & 5.3 & 6.4 & 13.9 & 14.7 & 9.4 & 6.7 & 8.9 \\
        & \scriptsize{MAST3R-SLAM\cite{murai2025mast3r}} & 5.3 & \textbf{2.5} & \textbf{1.5} & 9.7 & \textbf{8.8} & \underline{4.1} & \textbf{1.1} & \textbf{4.7} \\        
        & \scriptsize{ViSTA-SLAM\cite{zhang2025vista}} & 7.3 & 3.5 & 2.8 & \textbf{5.5} & 12.9 & \textbf{3.5} & 2.9 & 5.5 \\
        & \scriptsize{VGGT-Long\textsuperscript{$\dagger$}\cite{deng2025vggt}} & 80.2 & 82.3 & 49.5 & 65.6 & 68.2 & 48.1 & 84.4 & 68.3 \\
        & \scriptsize{LoGeR*\textsuperscript{$\dagger$}\cite{zhang2026loger}} & 8.6 & 4.1 & 4.8 & 9.7 & 15.0 & 7.7 & 2.8 & 7.5 \\
        & \scriptsize{Lingbot-map\cite{chen2026lingbot}} & 9.4 & 14.6 & 11.8 & 12.7 & 16.0 & 13.4 & 7.9 & 12.3 \\
        & \cellcolor{gray!15}\scriptsize{Ours (w/o BA)} & \cellcolor{gray!15}6.8 & \cellcolor{gray!15}6.0 & \cellcolor{gray!15}7.7 & \cellcolor{gray!15}8.8 & \cellcolor{gray!15}18.9 & \cellcolor{gray!15}6.8 & \cellcolor{gray!15}5.5 & \cellcolor{gray!15}8.6 \\
        \hhline{----------}
        \multirow{2}{*}{\begin{sideways}SE(3)\end{sideways}}
        & \scriptsize{DROID+M3D\textsuperscript{$\dagger$}\cite{teed2021droid,hu2024metric3d}} & \underline{9.3} & \underline{6.5} & \textbf{19.9} & \underline{16.6} & \underline{21.0} & \textbf{5.0} & \underline{16.2} & \underline{13.5} \\
        & \cellcolor{gray!15}\scriptsize{Ours (w/o BA)} & \cellcolor{gray!15} \textbf{7.1}& \cellcolor{gray!15} \textbf{6.3}& \cellcolor{gray!15} \underline{21.0}& \cellcolor{gray!15} \textbf{15.3}& \cellcolor{gray!15} \textbf{19.6}& \cellcolor{gray!15} \underline{6.8}& \cellcolor{gray!15} \textbf{8.2}& \cellcolor{gray!15} \textbf{12.0}\\
        \bottomrule
    \end{tabularx}
    \renewcommand{\arraystretch}{1.0}
    \vspace*{1pt}
    \raggedright
    
    \scriptsize{
    \textbf{pump.}\ and \textbf{kit.}\ are the abbreviations of pumpkin and kitchen.
    Methods with \textsuperscript{$\dagger$} are run by ourselves.
    }
    \label{tab:7scenes_traj_eval}
\end{table}

\subsection{Ablation study}
\label{sec:experiment_ablation}

To demonstrate the effectiveness of our tightly coupled scale formulation, we conduct ablations on the scale-aware optimization pipeline. 
Specifically, we first remove the scale optimization while retaining our proposed decoupled uncertainty modeling, and then assess the impact of the uncertainty decoupling by replacing it with a naive per-pixel uncertainty formulation.

For the first ablation, we remove scale estimation from both the frontend and backend, equivalent to treating scale as a constant $s=1$ in all states, marked as \textit{w/o s-opt}. 
In this setting, the scale error term in~\cref{eq:scale_error} is set to zero, and the depth error term in~\cref{eq:depth_error} relies solely on the metric depth predicted by the network to optimize landmarks and poses.
Metric depth uncertainty is modeled by propagating decoupled uncertainties through our proposed framework via~\cref{eq:uncertainty_propagation}, marked as \textit{prop.} in \cref{tab:kitti_ablation}.
Comparing \textit{Scalix} and \textit{w/o s-opt (prop.)} columns, using the same decoupled network but different SLAM backend integrations, our method improves \ac{ate} under both SE(3) and Sim(3) evaluation, indicating that scale-aware backend optimization enhances accuracy and consistency.

Under the configuration without scale optimization, we further compare our uncertainty propagation strategy against directly learning per-pixel independent uncertainty from the metric depth predictions (\textit{direct}).
Comparing \textit{prop.} and \textit{direct} columns in~\cref{tab:kitti_ablation}, incorporating scale-dependent correlations leads to similar SLAM performance, demonstrating that our training strategy does not degrade the uncertainty accuracy.

\begin{table}[]
    \centering
    \caption{Trajectory evaluation on the Kitti dataset. RMSE ATE [m] with SIM(3) and SE(3) alignments (\textbf{Best}, and \underline{second best}).}
    \setlength{\tabcolsep}{2pt}
    \renewcommand{\arraystretch}{1.1}
    \footnotesize
    \def\failcolor{gray}
    \begin{tabularx}{\columnwidth}{c|YYY|YYY}
        \toprule
        & \multicolumn{3}{c|}{SIM(3)} & \multicolumn{3}{c}{SE(3)}
        \\
        \hhline{~|---|---}
        & \multirow{2}{*}{\textbf{Scalix}} & \multicolumn{2}{c|}{\textbf{w/o s-opt}} & \multirow{2}{*}{\textbf{Scalix}} & \multicolumn{2}{c}{\textbf{w/o s-opt}}
        \\
        \hhline{~|~--|~--}
        \textbf{Seq.} & & \textbf{prop.} & \textbf{direct} & & \textbf{prop.} & \textbf{direct}
        \\
        \hhline{-|---|---}
        \textbf{00} & 2.85 & \underline{2.56} & \textbf{2.17} & 3.96 & \underline{3.55} & \textbf{2.18}
        \\
        \textbf{01} & \textbf{10.72} & \textcolor{\failcolor}{96.91} & \underline{66.65} & \textbf{21.47} & \textcolor{\failcolor}{\underline{575.58}} & \textcolor{\failcolor}{618.65}
        \\
        \textbf{02} & \textbf{15.04} & 18.23 & \underline{18.16} & \textbf{16.55} & 20.11 & \underline{18.47}
        \\
        \textbf{03} & 1.90 & \underline{1.84} & \textbf{1.67} & \textbf{9.44} & \underline{10.76} & 13.47
        \\
        \textbf{04} & \textbf{1.04} & \underline{1.28} & 1.86 & \textbf{3.25} & \underline{4.03} & 8.83
        \\
        \textbf{05} & 1.84 & \underline{1.72} & \textbf{1.57} & \underline{4.30} & \textbf{3.94} & 2.01
        \\
        \textbf{06} & \textbf{1.62} & \underline{1.66} & 2.21 & \underline{1.99} & \textbf{1.78} & 2.34
        \\
        \textbf{07} & 0.91 & \underline{0.87} & \textbf{0.71} & 2.15 & \underline{1.92} & \textbf{1.22}
        \\
        \textbf{08} & \textbf{7.30} & \underline{8.16} & 8.78 & \textbf{7.66} & \underline{8.16} & 9.25
        \\
        \textbf{09} & \textbf{4.76} & \underline{4.89} & 5.49 & \textbf{4.86} & \underline{5.56} & 9.33
        \\
        \textbf{10} & \underline{2.57} & \textbf{2.52} & 2.74 & \underline{3.46} & \textbf{2.77} & 3.70
        \\
        \hhline{-|---|---}
        \textbf{Avg.} & \textbf{4.60} & \textcolor{\failcolor}{12.78} & \underline{10.18} & \textbf{7.19} & \textcolor{\failcolor}{\underline{58.01}} & \textcolor{\failcolor}{62.67}
        \\
        % \textbf{Avg. success} & \underline{4.60} & \textcolor{\failcolor}{\textbf{4.38}} & 10.18 & 7.19 & \textcolor{\failcolor}{\textbf{6.27}} & \textcolor{\failcolor}{\underline{7.08}}
        %\\
        \bottomrule
    \end{tabularx}
    \renewcommand{\arraystretch}{1.0}
    \vspace*{1pt}
    \raggedright
    
    \scriptsize{
    \textbf{w/o s-opt} = without scale optimization. 
    \textbf{prop.} = propagated total uncertainty. 
    \textbf{direct} = directly-predicted total uncertainty.
    \textcolor{\failcolor}{Gray}: failures (ATE $>$ 3\% of seq. length).
    %\textbf{Avg. success}: average excluding failed sequences. 
    }
    \vspace{-0.2cm}
    \label{tab:kitti_ablation}
\end{table}

\section{Conclusion} 
\label{sec:conclusion}

In this paper, we have presented Scalix, a monocular SLAM system that achieves metric-scale trajectories for both indoor and outdoor environments. 
This is done by treating the scale as a measurement that can be optimized in a tightly-coupled fashion using a factor-graph formulation. 
To achieve this, we have proposed a novel parametrization and training strategy that allows foundation depth networks to learn scale and depth uncertainties, capturing the information and correlations of all the pixels used for monocular depth prediction. 
We have evaluated Scalix in both indoor and outdoor environments, demonstrating that it can be generalized to different scenarios. 
In particular, we demonstrate superior performance in outdoor scenarios, where Scalix improves the accuracy of the state estimates by $48\%$. 
In the future, we would like to investigate if this scale optimization strategy can be extended to multi-scale hypotheses while still using a tightly coupled and probabilistic formulation. 
Building on the factor-graph formulation, future work includes investigating how the proposed optimization integrates with multi-modal data from different embodiments, and how this can improve the robustness of the overall system.

%% Use plainnat to work nicely with natbib. 
\bibliographystyle{abbrvunsrt}
\bibliography{references}

\end{document}